\documentclass[]{interact}

\usepackage{amssymb}
\usepackage[labelsep=space]{caption}
\usepackage{color,soul}
\usepackage{natbib}
\usepackage[labelsep=space]{subcaption}
\usepackage[normalem]{ulem}
\usepackage{url}
\usepackage{xcolor}

\newcommand{\ie}{\textit{i.e., }}
\newcommand{\eg}{\textit{e.g., }}

\newcommand{\Task}{\textit{Task}}
\newcommand{\Action}{\textit{Action}}

\newcommand{\Actions}{\textit{Actions}}

\newcommand{\ObjectType}{\textit{Object Type}}

\begin{document}

\title{Hiding task-oriented programming complexity: an industrial case study}

\author{
\name{Enrico Villagrossi\textsuperscript{a}\thanks{CONTACT Enrico Villagrossi. Email: enrico.villagrossi@stiima.cnr.it}, Michele Delledonne\textsuperscript{a,b}, Marco Faroni\textsuperscript{a}, Manuel Beschi\textsuperscript{a,b}, and Nicola Pedrocchi\textsuperscript{a}}
\affil{\textsuperscript{a}Institute of Intelligent Industrial Technologies and Systems for Advanced Manufacturing, National Research Council of Italy, Via A. Corti 12, 20133, Milan, Italy; \\ \textsuperscript{b}Department of Mechanical and Industrial Engineering, University of Brescia, Via Branze 39, 25123 Brescia, Italy}
}
\thanks{\\\textbf{This article has been accepted for publication in the International Journal of Computer Integrated Manufacturing, published by Taylor \& Francis.}}

\maketitle

\begin{abstract}
The ease of use of robot programming interfaces represents a barrier to robot adoption in several manufacturing sectors because of  {the need for more expertise from the end-users}. Current robot programming methods are mostly the past heritage, with robot programmers reluctant to adopt new programming paradigms.  
This work aims to evaluate the impact on non-expert users of introducing a new \textit{task-oriented} programming interface that hides the complexity of a programming framework based on ROS. The paper compares the programming performance of such an interface with a classic \textit{robot-oriented} programming method based on a state-of-the-art robot teach pendant. 
An experimental campaign involved 22 non-expert users working on the programming of two industrial tasks.
 {\textit{Task-oriented} and \textit{robot-oriented} programming showed comparable  learning time, programming time and the number of questions raised during the programming phases, highlighting the possibility of a smooth introduction to \textit{task-oriented} programming even to non-expert users.}

\end{abstract}


\section{Introduction}\label{sec:introduction}
\subsection{Context}\label{ssec:context}
Industrial robots in manufacturing have steadily grown in the last few decades. According to the World Robotics Industrial Robot report 2021~\citep{IFR:2021}, two of the most addressed applications of industrial robots are material handling and components assembly, which are mainly pick\&place tasks.
The structure of a pick\&place program is frequently defined as a long list of \textit{``move to''} instructions. 

Since early robotic applications, the definition of the robot program was based on a \textit{teaching-by-showing} approach, where the programmer guides the robot manually along the desired trajectory (\textit{lead-through} programming), or with the Teach Pendant (\textit{drive-through} programming). 
 {The use of \textit{lead-through} programming is frequent for collaborative robots. The lightweight structure and the low payload allow for dragging the end-effector intuitively across the workspace, while the design of the robot allows for a safe interaction following the current safety standards as the ISO/TS 15066:2016~\citep{ISOTS15066:2016}.}

The evolution of robot programming techniques brought \textit{robot-oriented} programming languages~\citep{Yang:2015}. They are high-level programming languages, primarily based on BASIC and PASCAL, such as the ABB Rapid, the Fanuc Karel and the Kuka KRL. Such programming languages are integrated with advanced robotic functions that allow the development of complex robotic applications but retain the \textit{teaching-by-showing} programming mode. However, the languages continued to develop, gradually incorporating features from the rest of the programming world.
As a drawback, each robot manufacturer developed its proprietary \textit{robot-oriented} programming language incompatible with the others.
Reaching an essential knowledge of proprietary languages requires a reduced amount of time.  {However, a deep knowledge of a robotic platform requires great experience and training, which is why robotic programmers tend to become specialised in a few specific robot brands. This approach may cause resistance by robot programmers and robotic system integrators to acquire new programming skills, learn different programming paradigms, and use different robot brands.}  

The \textit{robot-oriented} programming languages are currently the most widespread for industrial applications, even if offline programming environments more and more often flank them.
Thanks to accurate robotic cell modelling, the development and the simulation of the robot programs, with offline tools, before on-site testing allows saving time~\citep{Pan:2012}.  {Integrating offline programming environments with CAD systems has also led to the automatic generation of the robot part program ~\citep{Castro:2019}.} Classic use is for continuous processes such as machining, painting and arc welding applications (\ie a vast number of via-points are generated by a CAD/CAM system and interpolated by the robot).
Offline programming can partially generate collision-free trajectories for a given planning environment.
More frequently, they are used to check and highlight the presence of collisions between the robot and the environment.
As for \textit{robot-oriented} programming languages, every robot manufacturer developed its offline programming environment, such as ABB RobotStudio, Kuka.Sim, Motoman MotSim, Fanuc RoboGuide. 
Nevertheless, an accurate 3D model of the cell is not always available, and the construction of the cell can bring inaccuracies w.r.t. the original design.
Hence, adopting the traditional \textit{drive-through} programming is frequent during the commissioning phase to re-teach trajectories via-points.

Despite (i) the increasing complexity of robotic applications, (ii) the evolution of \textit{robot-oriented} programming languages, (iii) the adoption of offline programming tools, the approach to robot programming and how the robot programs are structured have barely changed over the years. The robot program maintains a rigid structure as a list of instructions coded and saved into the robot memory.
Moreover, \textit{robot-oriented} programming languages and offline programming tools are not easily handled by end-users unfamiliar with robotic knowledge.

The introduction of the ROS framework in robotic research has boosted the search for a programming approach oriented to standardisation enabling the technological transfer of state-of-the-art algorithms from research to industry. ROS adoption can bring several innovations, such as \textit{task-oriented} programming, state-of-the-art motion planners, dynamic motion planning, simulation environments, and easy sensor integration. 
Unfortunately, ROS does not penetrate the world of industrial applications despite the efforts made by the ROS-industrial consortium~\citep{ROS-I:2022}. Currently, ROS-industrial is moving on ROS2 with a new tentative to relaunch the ROS initiative dedicated to industrial purposes.
The main barrier to ROS adoption in the industry remains the slow learning rate for the robotic technicians unfamiliar with General Purpose Languages (GPL), such as \textit{C/C++}, \textit{Java} and \textit{Python}, and the concepts of a robotic framework. 
Improvements to programming interfaces are required to attract new users unfamiliar with GPL bringing together the advanced features provided by ROS with the ease of use of classical \textit{robot-oriented} programming languages and a design that enables \textit{task-oriented} programming.

\subsection{Motivation and Contribution}\label{ssec:contribution}
Despite several demonstrations in robotic research, the penetration of advanced robot programming techniques, such as visual programming or programming by demonstration, is facing barriers in the industrial context~\citep{Villani:2018}. The main obstacles are the robustness of the advanced programming algorithms, the complexity of the programming interfaces and the technical heritage of robot programmers reluctant to adopt new programming techniques because of their familiarity with textual \textit{robot-oriented} programming languages.

 {The Manipulation Framework was introduced (MF), as described in~\citep{Villagrossi:2021}}, to evolve the current industrial robot programming paradigm moving from a \textit{robot-oriented} paradigm to a \textit{task-oriented} programming tool, to reduce the programming time and to simplify the development of complex collaborative applications. The MF improved with the introduction of a \textit{task-oriented} Graphical User Interface (MFI hereafter) designed to hide the complexity of the framework based on ROS.
This paper aims to compare the acceptance level, the ease of use and the effectiveness of the \textit{task-oriented} interface to program industrial tasks by non-expert robot programmers and end-users.
The paper compares the tasks programming made with the MFI  and a classic \textit{lead-through} programming approach made with the robot Teach Pendant (TP). The setup used for the test was a UR10e robot, with its TP, which is nowadays considered the state-of-the-art of robot TPs in terms of intuitiveness and ease of use\footnote{Standard industrial robot TPs are, in general, much more complex than UR's TP.}.

The study involved a heterogeneous group of university students from multiple STEM faculties and machine tool operators (with different backgrounds) as end-users testers. Only a few testers had little prior experience in robot programming, and no one had seen either the MFI or the TP used for the experiments. 

The experimental results demonstrate that using a \textit{task-oriented} framework, as the MFI, can introduce slightly longer learning time but can bring several benefits w.r.t. a standard programming technique. 
Advantages are evident when it is necessary to deal with: task repetition, robot reprogramming and collision-free motion planning.
%
%
The parameters monitored during the experimental tests to compare the interfaces are the learning time, the programming time, the number of questions made during the programming, the testing time and the reprogramming time. 

 {The comparison between the two programming methods, the results of an experimental campaign involving industrial tasks in a real industrial environment and the presence of machine tools operators in the study (which is not common in previous works) constitute interesting points of novelty.}

\subsubsection{Paper outline}\label{ssec:paper_outline}
The paper is organised as follows: Section~\ref{sec:related_works} analyses the related works, Section~\ref{sec:materials_and_methods} describes the experimental setup, the experiments design, the programming interfaces and the method. Section~\ref{sec:res_and_disc} discusses the results obtained during the experimental campaign.
Finally, Section~\ref{sec:conc_and_fut_works} reports the conclusions and the future works.

\section{Related works}\label{sec:related_works}
%
Programming interfaces evolved with the improvements in the technology~\citep{Siciliano:2010} providing a wide range of solutions~\citep{Tsarouchi:2016, Mukherjee:2022, Heimann:2020}.
However, a gap remains to bring robot programming closer to end-users~\citep{Ajaykumar_2:2021}, such as production operators, and promote the spread of robotic applications to new manufacturing fields~\citep{Ajaykumar:2021}.
Advanced approaches, such as programming by demonstration, visual programming, augmented/virtual reality, and natural language programming, are characterised by high intuitiveness since they constitute instances of natural and tangible user interfaces (NUIs and TUIs).

Programming by demonstration is a technique where the programmer can demonstrate the task to the robot~\citep{Billard:2008, Zhang:2016}. A common approach to programming by demonstration is the evolution of \textit{lead-through} programming method exploiting admittance/impedance control algorithms where the user physically interacts with the robot through robot manual guidance {~\citep{Safeea:2022}}. 
The most common approach uses force/torque sensors mounted between the robot flange and the end-effector~\citep{Bascetta:2013}. Conversely, modern collaborative robots, such as the Kuka LBR or the Franka Emika, integrate torque sensors into robot joints. 
The drawbacks of admittance/impedance control algorithms are accurate parameters tuning, which can bring stability issues~\citep{Ferraguti:2017}, additional sensing (force/torque sensors), and they are not available as a software feature for most industrial robots. Moreover, the demonstration accuracy is not enough for most industrial applications.

Alternative methods, less common, provide the use of voice, vision {~\citep{Shuang:2022}} and motion capturing~\citep{Makris:2014}.

%
Visual programming makes programming more approachable for non-experts~\citep{Corornado:2020, Huang:2020}. In literature, visual programming interfaces commonly use flow diagrams, behaviour trees, blocks, and icons~\citep{Stenmark:2017}. Some visual programming systems propose personal graphical user interfaces~\citep{Casper:2018} that use buttons, menus, windows, textual inputs, and sliders. The standard IEC-61131-3, which defines programming languages for automation, provide visual programming languages, such as the Function Block (FB) or the Sequential Flow Chart (SFC), that can also be used for robot programming as standard languages~\citep{Thormann:2021, Rendiniello:2020}. 
These languages are suitable for beginner programmers. The execution speed
of visual applications is slow, and the programming requires more time than a textual one. A complex task requires a large number of operations, users spend time making
room for things, encasing and rearranging them in macros, and the overall program
can get crowded.

Augmented/virtual reality allows overlapping the real-world environment with a virtual one. With this technology, some information or programming tools can appear directly in the environment. The possibility of overlapping a virtual robot and objects allows the operator to use the programming by demonstration without interacting directly with the robot~\citep{Blankemeyer:2018}. Visualising virtual panels with programming information (\eg robot trajectory or parameter values) allows the operator to make decisions~\citep{Gadre:2019}. 
 {Using physical auxiliary tools~\citep{Ong:2020} or object detection software~\citep{Apostolopoulos:2022} can facilitate the user programming.} 
This technology presents a high implementation cost; it needs to be more flexible as it is programmed for a specific task and prone to failures in case of environmental changes. 

Natural language programming uses speech and text to create programs.
Usually, this technique works in parallel with another programming technique. The high complexity of human language requires some constraints, and this type of language cannot describe the action in its entirety. 
In literature, natural language programming was combined with programming by demonstration~\citep{Quintero:2018}, visual programming~\citep{Huang:2017}, and augmented reality~\citep{Andronas:2021}. The use of speech recognition is hardly usable in industrial environments; the high noise that characterises these working places makes it difficult to recognise voice commands.

Tangible programming is a technique that uses physical objects to define program structures. 
By positioning specific objects in the environments, it is possible to define the desired object for a particular action, the action to perform, and the areas where the actors perform it~\citep{Sefidgar:2017}. Another approach is to use the specific card to define the task structure; the card type and order represent the action types and order~\citep{Kubota:2020}. The programming of a complex task might require a large number of objects making the environment chaotic. Some actions are difficult to describe through objects, especially if they need good accuracy.

Alternative ways to automate the creation of the robot part-program provide the direct use of a CAD model to extract the necessary information to control the robot, bypassing the robotic offline programming tools as in~\citep{Neto:2013}. This approach is helpful for continuous processes.

The EU projects Robo-Partner~\citep{RoboPartner:2016}, Sharework~\citep{Sharework:2022} and Sherlock~\citep{Sherlock:2022}, within their scopes, investigated the use of advanced programming techniques in industrial scenarios. 
Robo-Partner faced the programming of an assembly task exploiting programming by demonstration techniques through audio commands, visual programming, and direct robot arm manipulation by the user through force/tactile sensing, as described in~\citep{Michalos:2014}. The goal of Robo-Partner was to bring robotics closer to SMEs where poor robotic knowledge represents a barrier to robot adoption. 
Sharework aims to introduce dynamic task planning able to assign the robot actions to the robot controllers autonomously {~\citep{Umbrico:2022}}. The assignment evaluates the evolution of a collaborative task controlling the robot through ROS libraries~\citep{Faroni:2020}. The project developed robot motion planning algorithms based on learning by demonstration and provided intuitive human-robot interaction methods based on virtual reality.
Sherlock aims to provide zero-programming robotics for collaborative and medium-high payload robots. The project developed a programming architecture that mixes augmented reality (\ie indirect interaction) and manual guidance (\ie direct interaction) based on force sensors. Mixed reality roughly defines the trajectory starting and ending point, while manual guidance can refine and adjust precisely the final position. Using a motion planner and a digital twin enables the generations of collision-free trajectories~\citep{Sherlock_D4-8:2022}.

Several frameworks such as: CORBA~\citep{Schmidt:1998}, YARP~\citep{Metta:2006}, ROS~\citep{ROS:2022}, IMI2S~\citep{Anzalone:2014} are currently used to control the robot. Adopting frameworks is a prerogative of robotic research; no relevant spread is currently perceptible in the industry.
Among the others, the most successful is ROS, representing a standard \textit{de facto} for robotic research programming. ROS brings advanced features and open-source packages that allow for a \textit{task-oriented} approach.
On the other hand, ROS imposes the use of GPL languages, which are often adopted only in research\footnote{Kuka Sunrise.OS~\citep{SunriseOS:2022}, dedicated only to Kuka LBR iiwa, is currently a unique example of a robotic OS that provides libraries based on GPL languages, like Java, to program the robot.}.  

\section{Materials and Methods}\label{sec:materials_and_methods}

\subsection{The programming interfaces}\label{ssec:progr_intf}
\begin{figure}[tpb]
    \centering
    \includegraphics[width=0.5\columnwidth]{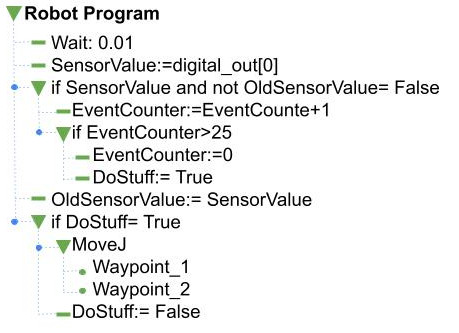}
    \caption{Robot-oriented program.}
    \label{fig:robot_oriented}
\end{figure}

This paper compares a \emph{robot-oriented} and a \emph{task-oriented} programming interface.
\textit{Robot-oriented} programming focuses on primitive robot movements that the robot can perform.
The user combines these primitive actions into a sequence to obtain the desired program. 
Figure~\ref{fig:robot_oriented} shows an example of a robot-oriented program.

\begin{figure}[tpb]
    \centering
    \includegraphics[width=0.5\columnwidth]{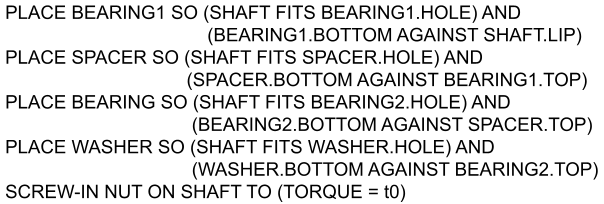}
    \caption{Task-oriented program.}
    \label{fig:task_oriented}
\end{figure}

\textit{Task-oriented} programming focuses on the task. The user combines high-level actions by setting the parameters required by the process operation rather than the robot's motion.
The user does not define the primitive action from scratch, as the framework programmer previously defined the task structure. 
The user codes in an intuitive language, as shown in Figure \ref{fig:task_oriented}.

The experiments involved one programming interface for each type. The \textit{robot-oriented} programming interface is the UR10e TP, and the \textit{task-oriented} one is a novel GUI built on the top of the Manipulation Framework~\citep{Villagrossi:2021}.

\subsubsection{Robot-oriented programming interface: UR10e teach pendant interface}\label{sssec:UR10ei}

\begin{figure}[t]
    \centering
    \includegraphics[width=0.7\columnwidth]{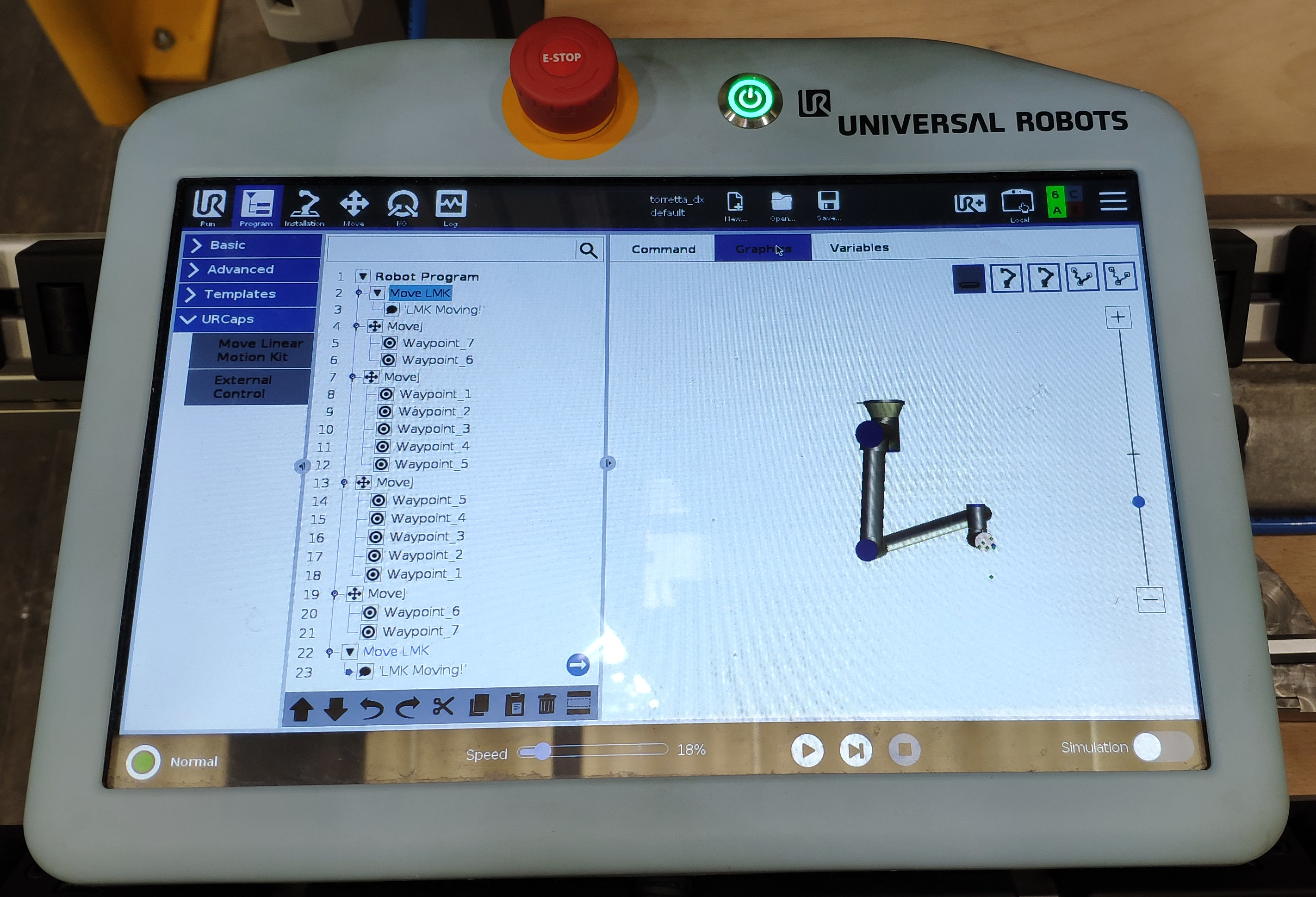}
    \caption{UR10e teach pendant.}
    \label{fig:UR10TP}
\end{figure}

The UR10e TP is a highly intuitive programming interface based on a touch screen device without physical buttons (apart from the on/off and the emergency buttons); Figure~\ref{fig:UR10TP} shows the UR10e TP. The TP enables the robot movements in jogging mode, the access to robot configurations and parameters, and the robot programming through a \textit{robot-oriented} high-level programming language.
A new robot program requires a sequence of \textit{move} instructions by teaching the starting and the ending robot configuration to be interpolated. The teaching of robot position can be done by \textit{lead-through} programming by moving the robot with the so-called manual guidance mode.  {The programmer must add intermediate robot configurations (via-points) to guarantee collision-free trajectories.}
The TP interface (TPI) provides specific functions to manage the gripper activation.

\subsubsection{Task-oriented programming interface: manipulation framework interface}\label{sssec:MFi}
\begin{figure}[t]
    \centering
    \includegraphics[width=0.9\columnwidth]{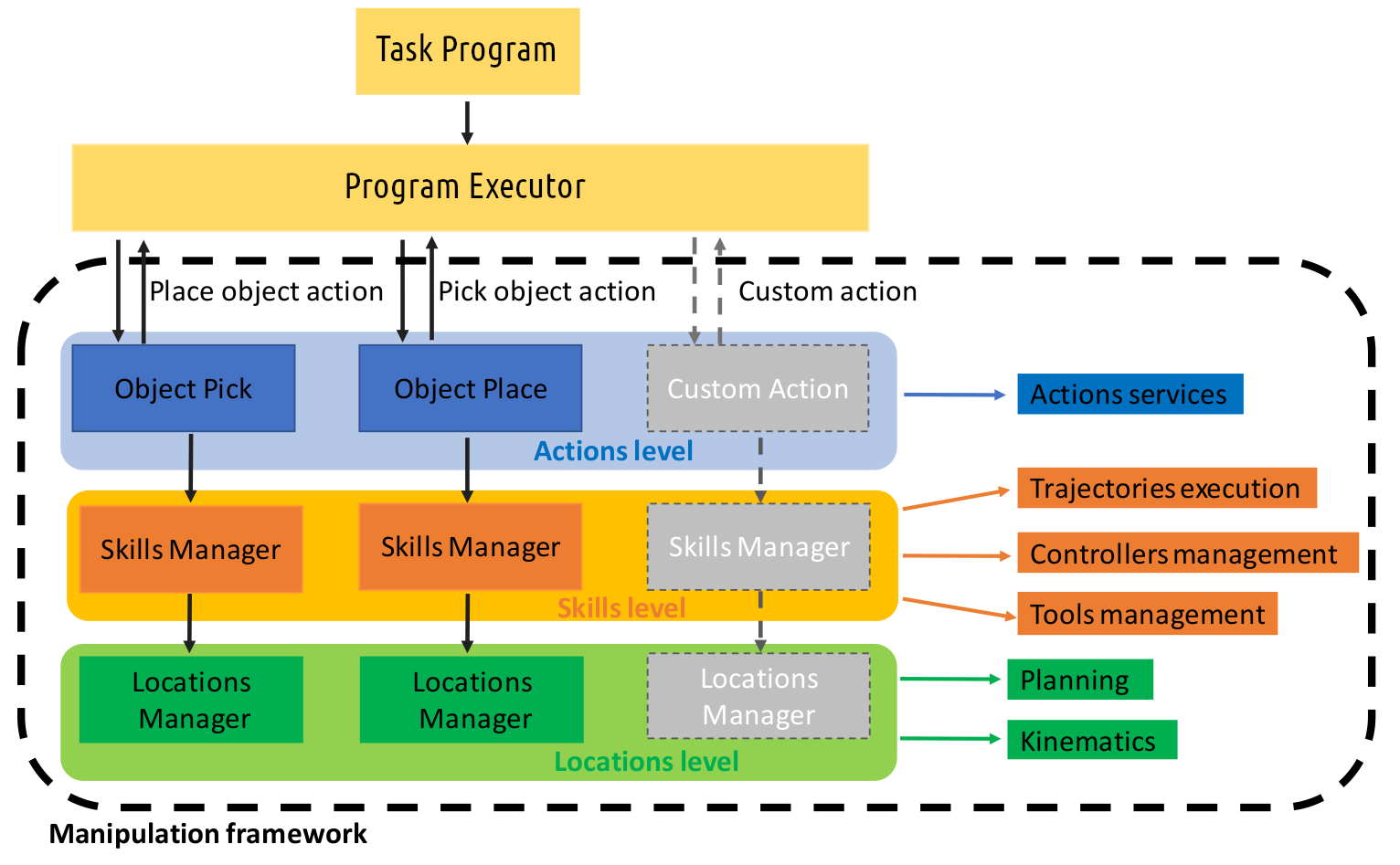}
    \caption{Manipulation framework layers description.}
    \label{fig:manipulation_framework}
\end{figure}

The MF is a software package dedicated to manipulation tasks~\citep{Villagrossi:manipulation-examples-github,Villagrossi:manipulation-github}. 
MF provides some essential components to enable \textit{task-oriented} robot programming.
Figure~\ref{fig:manipulation_framework} shows the structure of the MF.
For a given manipulation task, the MF aims to: (i) process manipulation actions (such as pick, place, move to), relieving the programmer from the management and the execution of single actions;
(ii) handle the kinematic model of the robotic system (arm + grasping system) being able to compute forward and inverse kinematics; (iii) embed motion planning functionalities to generate
collision-free trajectories for a given planning scene and a given robotic system in the planning environment that can dynamically change; (iv) execute the planned trajectories on the desired robotic system; (v) change the robot controller dynamically. 
 {The motion planning uses \textit{MoveIt!} pipeline~\citep{Moveit:2022}. \textit{MoveIt!} allows dynamic load motion planners as plugins, enabling collision-free motion planning for a given planning scene. Additional information is available in~\citep{Villagrossi:2021}.}

\begin{figure}
     \centering  
     \includegraphics[width=0.9\textwidth]{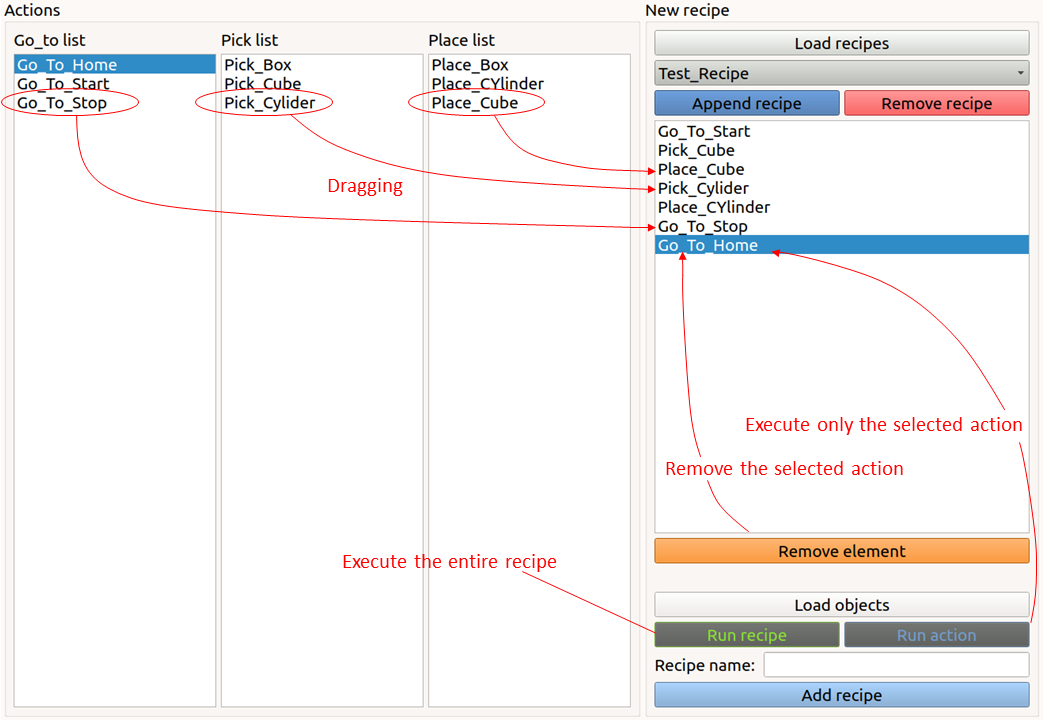}
     \caption{MFI: example of \Task\ composition by concatenating multiple \Actions.}
     \label{fig:recipe}
\end{figure}

\begin{figure*}[tpb]
	\centering
	\subfloat[][Action setting]
	{\includegraphics[height=0.7\textwidth]{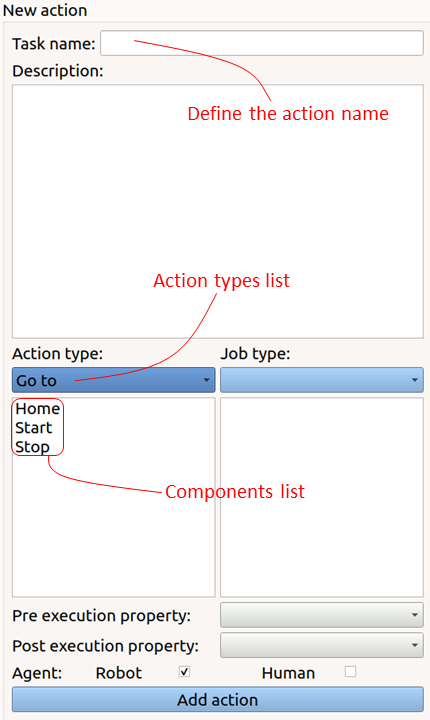}
	\label{fig:action_setting_win}}\qquad
	\subfloat[][Object setting]
	{\includegraphics[height=0.7\textwidth]{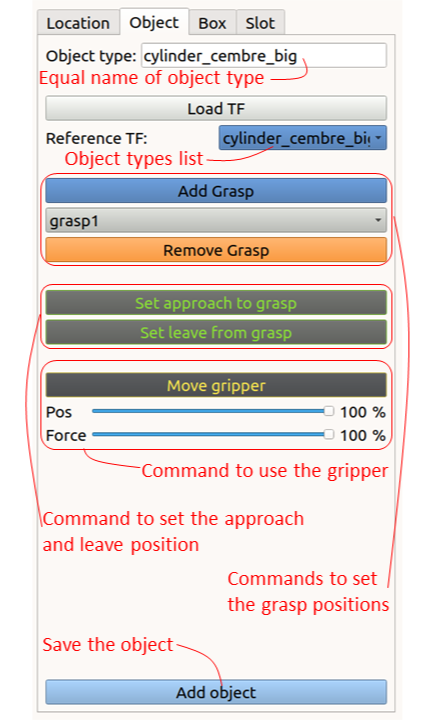}
	\label{fig:objects_win}}
	\caption{Action and components setting panels.}
	\label{fig:components_screenshot}
\end{figure*}

 {MF's APIs require \textit{C++} programming and ROS integration (\emph{e.g.}, the creation of ROS nodes, the definition of configuration files, and dataset management.)} To overcome this limit, a GUI was developed as MF Interface (MFI)~\citep{Delledonne:manipulation-interface-github, Delledonne:manipulation-interface-example-github}. 
MFI has a structure designed for \textit{task-oriented} programming. 
The interface allows to easily concatenate multiple \Actions\ to assemble a \Task.
The user can set all the components using interaction with the robot being free to focus on the task definition. 
The user can choose between pre-programmed \Actions, such as \textit{Pick, Place, GoTo}. 
A program can be composed by dragging and dropping the desired \Actions, as shown in Figure~\ref{fig:recipe}. 
A set of parameters defines each \Action. For example, a \emph{Pick} action requires the approaching pose, the grasping pose and the gripper closing force (see Figure~\ref{fig:objects_win}).
Notice that the MFI does not require to specify via-points, as the MF will automatically plan the trajectories based on the task specifications.

A relevant feature of the MFI allows specifying an \ObjectType\ that groups several objects. 
The MF will automatically compute the poses for the object grasping (and IK solution) for every object of the same \ObjectType\ during the motion planning. The object with the optimal path planning will be selected.

\begin{figure}
     \centering  
     \captionsetup{justification=centering}
     \includegraphics[width=0.9\textwidth]{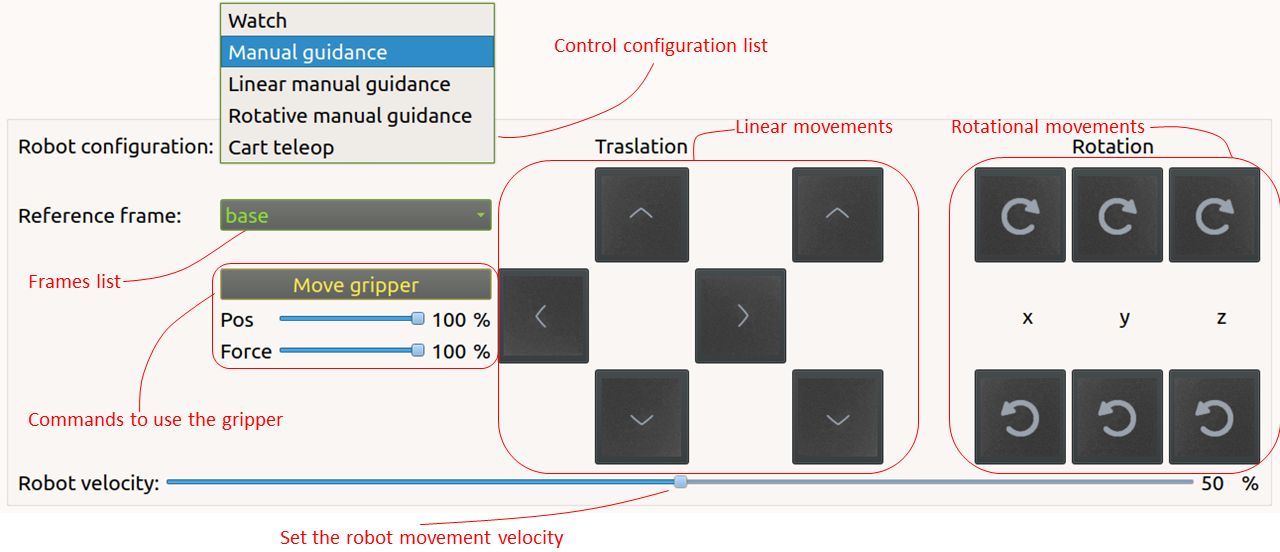}
     \caption{Interface Joystick, the movements are performed moving the end-effector frame with respect to the chosen reference frame.}
     \label{fig:joystick_screenshot}
\end{figure}

 {Positioning the robot in the desired position is possible to acquire a robot pose, similar to classic TP programming.}
The MFI allows for both \textit{lead-though} programming (if the robot has a force/torque sensor) and tele-operation, see Figure~\ref{fig:joystick_screenshot}.
The MFI can be easily connected to a vision system to automatically localise the objects' positions in the robot workspace, providing additional programming flexibility and autonomy. 

The interface allows for \Action, or sequence of \Actions\ execution. The first step is the kinematics computation (\ie robot and objects involved in the task), consequently, the computation of the collision-free motion plans. An error message informs the operator in case of unfeasible inverse kinematics or motion plan, so it is enabled the modification of the components or the \Action\ sequence to fix the issue.
 {Currently, the framework provides only three actions. The development of a new one still requires an expert developer; however, the number of actions needed to address complex industrial tasks is limited~\citep{Bogh:2012}, and the actions available in the MF allow it to cover a wide range of tasks. The possibility to intuitively generate a custom action, starting with the elementary skills, will be part of future works.}

\subsection{Experimental Setup}\label{ssec:setup}
The experimental setup consists of a Universal Robots UR10e mounted on a Cobotracks linear guide. The robot gripper is a Robotiq 2F-85. An external PC controls the robot at a frequency of $500 [Hz]$. 
The software used for the PC robot communication is the official ROS package~\citep{UniversalRobots:ROS-driver-github}. This package requires the micro-interpolated joint positions to control the robot. The feedback information are joint positions, velocities, currents and external forces.

\begin{figure}[t]
    \centering
    \includegraphics[width=0.8\columnwidth]{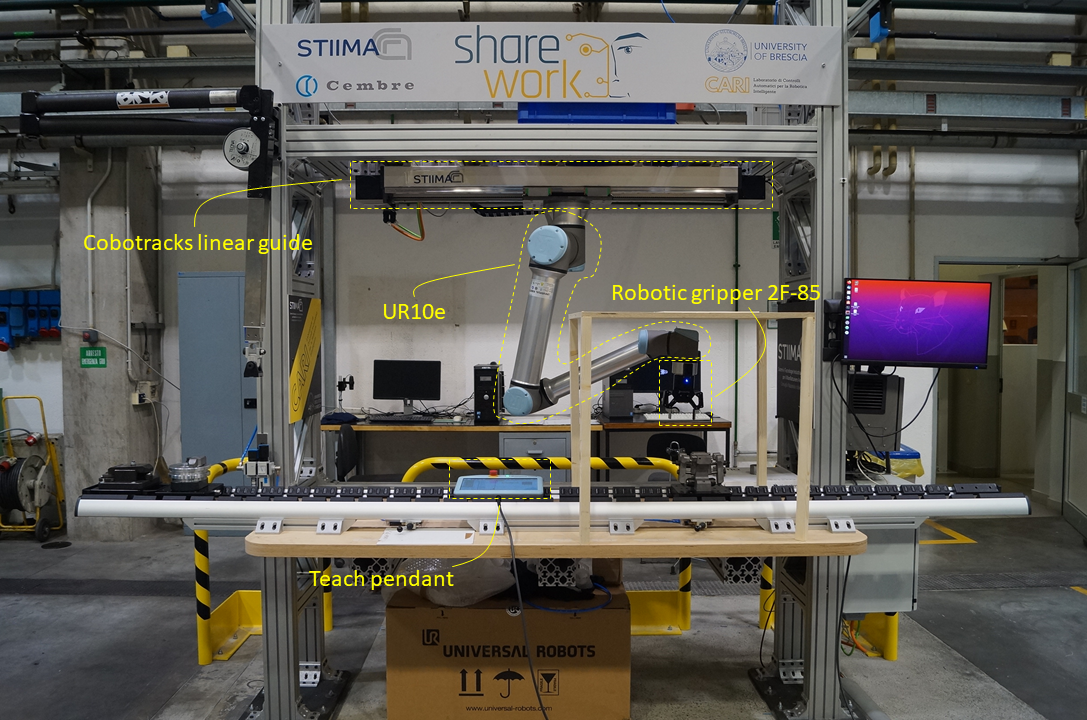}
    \caption{Experiment setup.}
    \label{fig:Setup}
\end{figure}

The robot's end-effector is a Robotiq gripper with two fingers and $85 [mm]$ of working range; a force/torque sensor was mounted between the robot flange and the gripper. The force measuring range is $100 [N]$, and the torque measuring range is $10 [Nm]$. 
 {Figure~\ref{fig:Setup} shows the experimental setup, highlighting the ceil robot configuration.} 
The Cobotracks linear guide was not used during the experiments.

\subsection{Experiments description}\label{ssec:experiments}

During the experimental session, the users programmed two tasks exploiting the \textit{robot-oriented}  {(\ie the UR10 Teach Pendant Interface, TPI hereafter)} and the \textit{task-oriented} (\ie the MFI) programming interfaces.
Task 1 was a pick\&place task with 10 objects, thought to measure the performance (\ie programming and testing time) of the two programming interfaces with a highly repetitive task. Task 2 recalls a machine tending application where the robot has to enter the machine tool workspace and withdraw or release a workpiece by avoiding obstacles. The task was thought to measure the performance of the interfaces when required to deal with multiple obstacles in the robot workspace, restricted spaces, and reprogramming.

The experiments consist of five phases:
\begin{enumerate}
    \item introduction: the user is informed about the experiment and the test phases. 
    \item Teaching: the user watches a video that describes the interfaces (\ie the robot TPI or the MFI) and their usage. 
    Then, an expert operator supports the user in assisted training, where the goal is to perform a single pick and place. 
    In this phase, the user is free to ask questions to the trainer. 
    The training continues until the user declares he/she can program a task autonomously. Finally, the expert operator describes the user's task to program; the programming phase can start. 
    \item Autonomous programming: the user programs the robot without the help of an expert. This phase ends when the user declares finished the task programming. The user can ask questions if he/she cannot proceed in task programming. The number of questions is an evaluation parameter.
    \item Testing: testing the program developed in the previous phase. In case the task is correctly performed, this phase ends. On the contrary, the user must correct and test the program until it is completed. The task's success determines the end of this phase. 
    \item Questionnaire: the user fills in a questionnaire regarding the intuitiveness and complexity of the interface.
\end{enumerate}

The experiments on Tasks 1 and 2 were carried out with 22 people. All subjects participated voluntarily, signing an informed consent form by the Declaration of Helsinki.

\subsubsection{Task 1 experiment description}\label{sssec:task1_exp_desc}
\begin{figure}
     \centering  
     \includegraphics[width=0.8\textwidth]{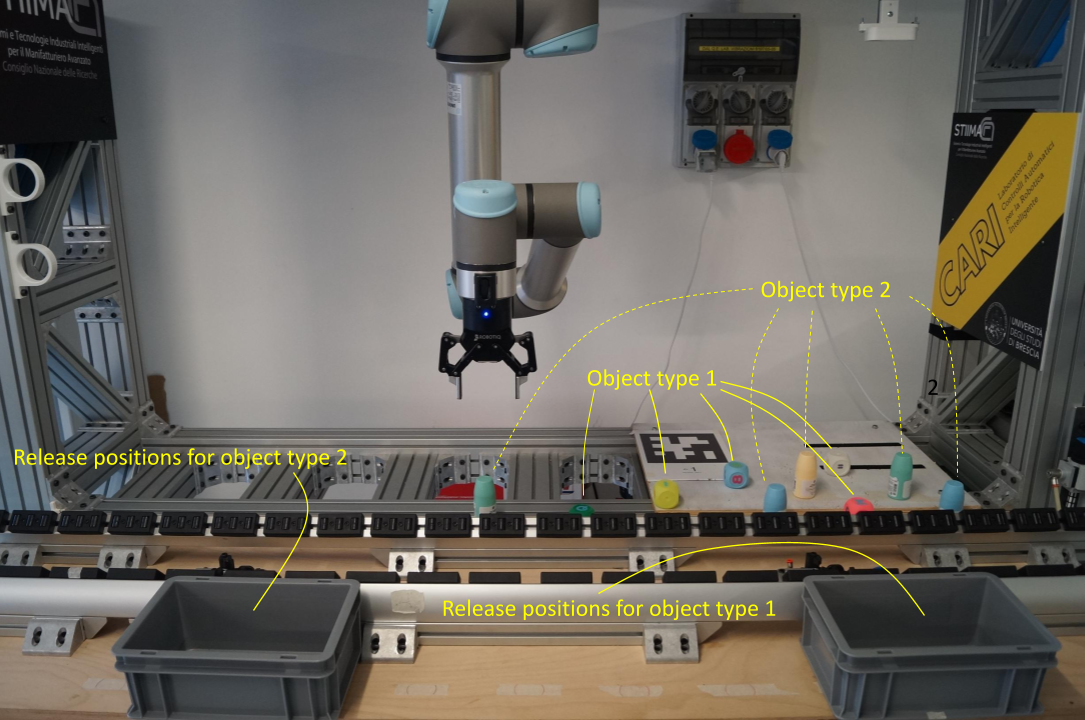}
     \caption{\textit{Task 1 setup and description.}}
     \label{fig:task1}
\end{figure}

Task1 requires the pick\&place of ten objects. The objects were of two types (cubes and cylinders), 5 objects per type.
After the picking, the robot place the object in the relative box on the base of the object type. The programmer is free to decide the picking sequence and, if necessary, trajectories via-points. Figure~\ref{fig:task1} shows the objects and boxes layout. The box on the right was for the cubes, and the box on the left was for the cylinders.

The experiments enlisted two groups of users, hereafter named Group A and Group B. 
Users were mostly university students from STEM faculties, aged between 20 and 35 years old. 
Group A was composed of 8 people; they programmed the application using the UR10e TP's described in Section~\ref{sssec:UR10ei}. 
Group B was composed of 9 people; they programmed the same application using the MFI described in Section~\ref{sssec:MFi}. 

The experiment consists of the five phases described in Section~\ref{ssec:experiments}. 

\subsubsection{Task 2 experiment description}\label{sssec:task2_exp_desc}

Task 2 requires picking 2 objects from a constrained box, moving them outside the box and placing them in a specific area. The picking sequence is free; the programmer can decide the picking positions. Once the program is finished and tested, the programmer has to change the release positions of the objects.

The idea is to avoid a penalisation of the TPI when a high number of repetitive \Actions\ is required, as for Task 1. Indeed, the task involves only 2 objects. At the same time, the robot has to move through a restricted environment with obstacles to evaluate the planning performance of the MF (\ie collision-free motion planning) compared to the TP where the programmer has to take care of the collision avoidance by adding intermediate trajectory via-points.

This task simulates a typical application of an industrial robot in machine tending, where the robot has to enter the machine tool workspace to withdraw or release a workpiece, and several obstacles constrain the planning environment. The programmer has to take care of possible collisions carefully. A frequent request is the reprogramming of robot tasks due to modifications to the robotic cell. Indeed, at the end of the task programming, we requested to adjust the robot program slightly to deal with robotic cell changes. We asked to partially modify the original robot program to measure the reprogramming time with both interfaces. These requests are frequent when small batches and huge production variability characterise the production.

The experiments' execution was on the shop floor of a mechanical engineering company.  
The experiment participants were machine tools operators and one production engineer with neither a robotic background nor previous experience in robot programming; it was the first time they had seen both programming interfaces.
The users' group was composed of 5 people, aged between 26 and 46 years old; 4 over 5 users have high school graduation, while the fifth has an automation engineering degree. The users were all with technical backgrounds and experiences in machine tool programming but no experience in industrial robot use.

The hosting company provided only a limited number of users, so it was impossible to create two independent groups; thus, all the participants repeated the experiments twice using both interfaces. The first experiment can influence the learning phase of the second one because the programming experience can provide some knowledge of the task for the second experiment; the order of the interfaces was randomly alternated, avoiding any possible bias.  {The users who experimented with Task 2 were not involved in Task 1 experiments' and vice-versa. }

%
This experiment was composed of the five phases described in Section~\ref{ssec:experiments}, with the addition of an autonomous reprogramming phase after phase 4 where the user has to modify the program already developed. The time spent by the user during the autonomous reprogramming and the number of questions were recorded as evaluation parameters. A second test phase has been added to correct any reprogramming errors.  {Finally, the user has to fill the questionnaire.}

\begin{figure}
     \centering  
     \includegraphics[width=0.8\textwidth]{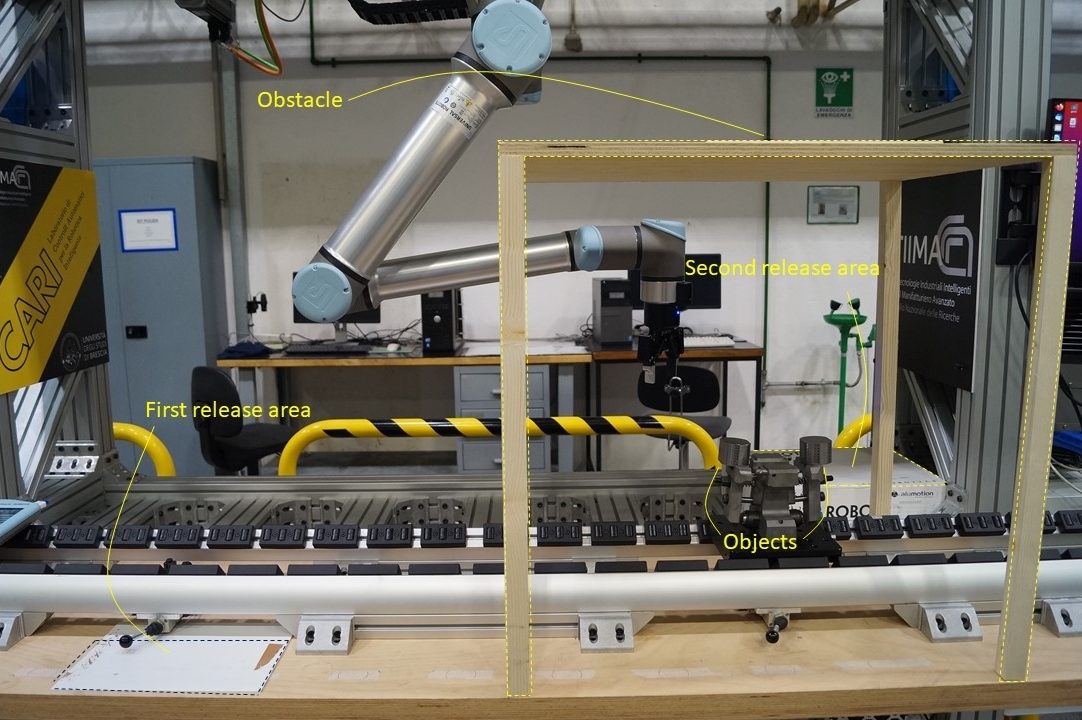}
     \caption{\textit{Task 2 setup and description.}}
     \label{fig:task2}
\end{figure}

\subsection{Performance Measurement}\label{ssec:progr_perf_meas}

In general, during the task programming, it is possible to identify 4 main phases: (i) the user learning phase, (ii) the program development phase, (iii) the testing phase, (iv) the execution phase. 
The first evaluation metric is the time taken by these phases. 
Therefore, the following evaluation parameters have been chosen:
\begin{itemize}
    \item learning time (\textbf{\emph{LeT}}): the time spent by the user to learn the programming environment during the assisted programming in phase 2 (teaching). \textit{LeT} evaluates the steepness of the learning curve of the interface.
    \item Programming time (\textbf{\emph{PrT}}): the time spent by the user to develop a robot program to achieve the given task. \textit{PrT} reflects the ease of use and the intuitiveness of the interface.
    \item Testing time (\textbf{\emph{TeT}}): the time spent by the user to test the robot program. The programmer must test the program developed in the real working environment by checking possible collisions and adding or adjusting via-points if needed. \textit{TeT} is a proxy of the ability of the interface to avoid user errors during task programming as it measures the time spent for program debugging.
    \item Execution time (\textbf{\emph{ExT}}): the time spent by the robot to execute the program. The execution time is related to the robot speed (equal for both the interfaces) and the length of the paths found by the motion planners. Using the TP, the \textit{ExT} depends on the number and the position of the via-points selected by the programmer; using the MFI the length of the trajectory is defined by the selected motion planner.
    \item Number of tests executed (\textbf{\emph{TeN}}): the number of executions run before a correct task execution (excluded \emph{ExT}). Similarly to \emph{TeT}, \textit{TeN} evaluates the robustness of the programming interface w.r.t. the errors made by the user during the programming.
    \item  {Number of questions during programming (\textbf{\emph{PrQ}}) and testing (\textbf{\emph{TeQ}}): the amount of questions highlights the user understanding of the interface use, it evaluate the ease of use and the intuitiveness of the interface;}   
\end{itemize}

During the experiments of \textit{Task 2}, the following parameters were also measured to evaluate the reprogramming phase:
\begin{itemize}
    \item reprogramming time (\textbf{\emph{ReT}}): the same as \emph{PrT} during the re-programming phase.
    \item reprogramming testing time (\textbf{\emph{ReTeT}}): the same as \emph{TeT} during the re-programming phase.
    \item reprogramming execution time (\textbf{\emph{ReExT}}): the same as \emph{ExT} during the re-programming phase.
    \item number of tests made during the reprogramming (\textbf{\emph{ReTeN}}): the same as \emph{TeN} during the re-programming phase.
    \item number of questions during reprogramming (\textbf{\emph{ReQ}}) and reprogramming testing (\textbf{\emph{ReTeQ}});  
\end{itemize}
Finally, to evaluate the user feel, users filled out the questionnaire in Table~\ref{tab:questionnaire}.

\begin{table}[tpb]
	\caption{Questionnaire proposed to users after Task 1 and Task 2.}
	\centering
	\footnotesize
        \begin{tabular}{cl} 
        \toprule
          1   & On a scale of 1 to 10, where 1 is inexperienced and 10 is very \\
              & experienced, how experienced are you in the use of industrial robots?   \\
          2   & On a scale of 1 to 10, where 1 is not simple and 10 is very simple, how \\
              & simple did you find the use of the programming interface you were proposed to use?   \\
          3   & On a scale of 1 to 10, where 1 is not intuitive and 10 is very intuitive, how\\
              & intuitive did you find the use of the programming interface you were proposed to use?   \\
          4   & On a scale of 1 to 10, where 1 is not fast and 10 is very fast, how fast did \\
              & you find learning to use the programming interface you were proposed to use?   \\
          5   & On a scale of 1 to 10, where 1 is not useful 10 is very useful,  how much did your previous\\
              & knowledge affect you in learning to use the programming interface you were proposed to use?   \\
          \bottomrule
        \end{tabular}
		\label{tab:questionnaire}
\end{table}
\section{Results and Discussion}\label{sec:res_and_disc}

The methodology applied during experiment execution is described in Section~\ref{ssec:experiments}.
The evaluation exploits the performance indexes described in Section~\ref{ssec:progr_perf_meas}.

\subsection{Task 1 Experiments Results}

\begin{figure}
     \centering
     \captionsetup{justification=centering}
     \begin{subfigure}[b]{0.46\textwidth}
         \centering
         \includegraphics[width=\textwidth]{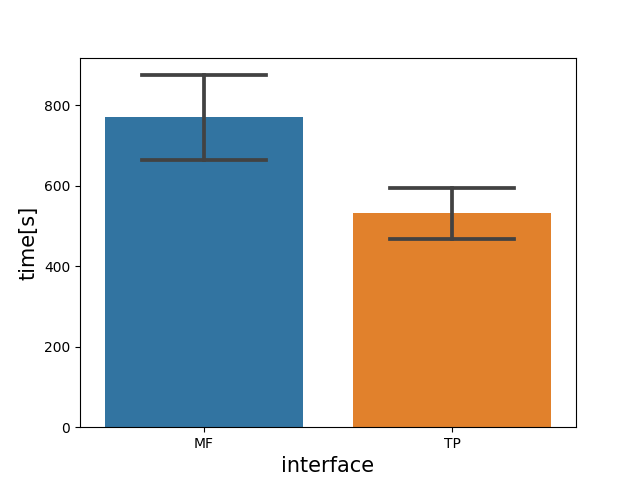}
         \caption{\textit{Learning Time (LeT)}.}
         \label{fig:Task1_LeT}
     \end{subfigure}
     \hfill
     \begin{subfigure}[b]{0.46\textwidth}
         \centering
         \includegraphics[width=\textwidth]{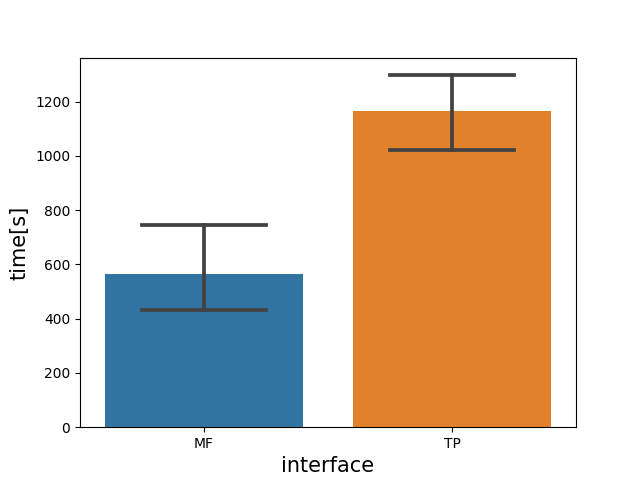}
         \caption{\textit{Programming Time (PrT).}}
         \label{fig:Task1_PrT}
     \end{subfigure}
     \hfill
     \begin{subfigure}[b]{0.46\textwidth}
         \centering
         \includegraphics[width=\textwidth]{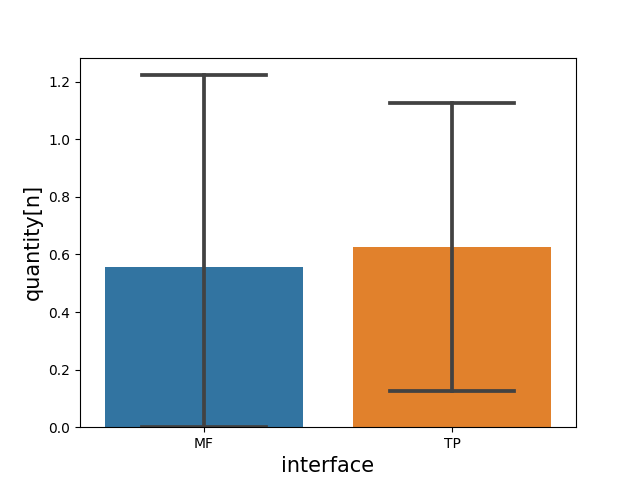}
         \caption{\textit{Programming Questions (PrQs).}}
         \label{fig:Task1_PrQ}
     \end{subfigure}
     \hfill
     \begin{subfigure}[b]{0.46\textwidth}
         \centering
         \includegraphics[width=\textwidth]{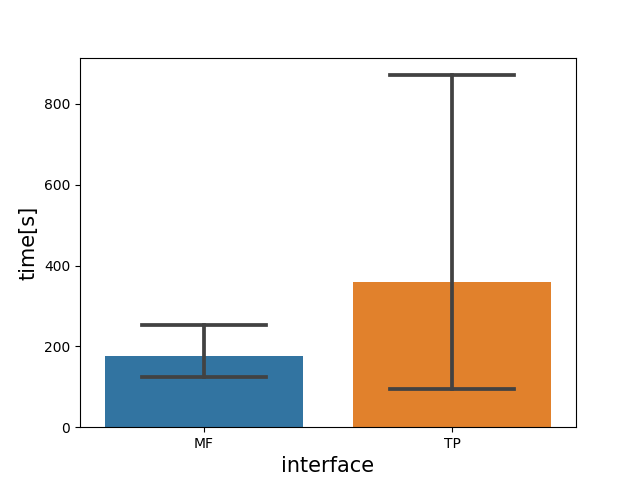}
         \caption{\textit{Test Time (TeT).}}
         \label{fig:Task1_TeT}
     \end{subfigure}
     \hfill
     \begin{subfigure}[b]{0.46\textwidth}
         \centering
         \includegraphics[width=\textwidth]{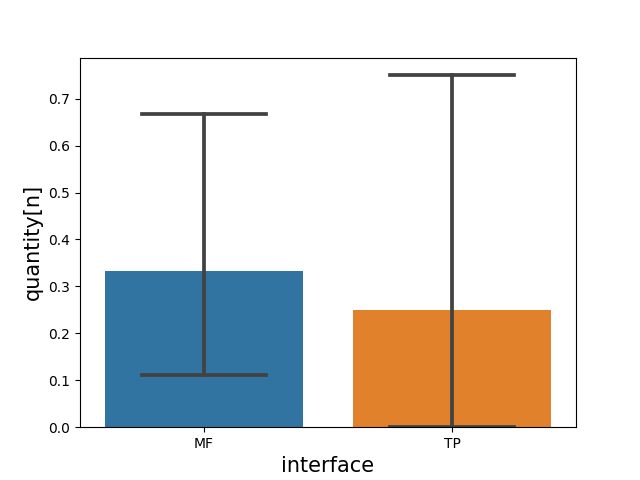}
         \caption{\textit{Test Numbers (TeNs).}}
         \label{fig:Task1_TeN}
     \end{subfigure}
     \hfill
     \begin{subfigure}[b]{0.46\textwidth}
         \centering
         \includegraphics[width=\textwidth]{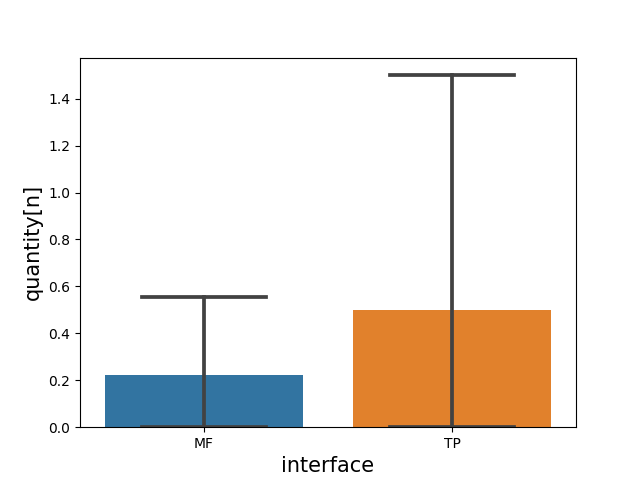}
         \caption{\textit{Test Questions (TeQs).}}
         \label{fig:Task1_TeQ}
     \end{subfigure}
     \hfill
     \begin{subfigure}[b]{0.46\textwidth}
         \centering
         \includegraphics[width=\textwidth]{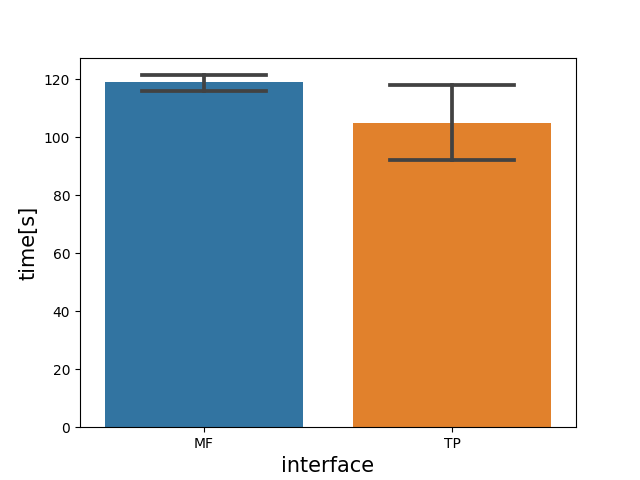}
         \caption{\textit{Execution Time (ExT).}}
         \label{fig:Task1_ExT}
     \end{subfigure}
     \hfill
     \begin{subfigure}[b]{0.46\textwidth}
         \centering
         \includegraphics[width=\textwidth]{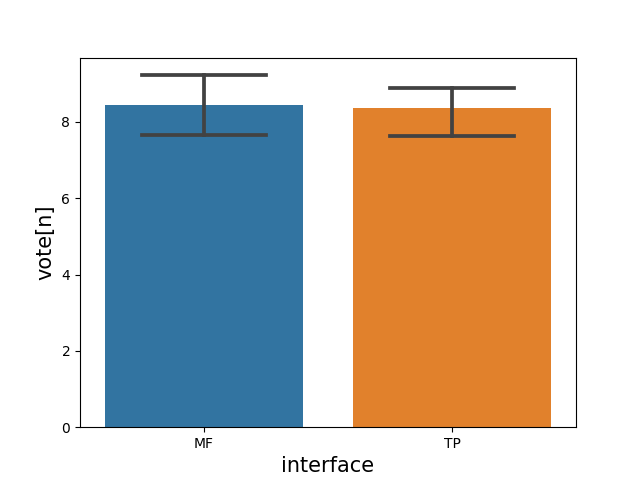}
         \caption{\textit{Interface ease of use, Table~\ref{tab:questionnaire} Q2.}}
         \label{fig:Task1_Comp}
     \end{subfigure}
        \caption{\textit{Task 1} experiments' results. \\ TP: UR10e teach pendant interface. MF: manipulation framework interface.}
        \label{fig:Task1_results_a}
\end{figure}

\begin{figure}
     \centering  
     \captionsetup{justification=centering}
     \begin{subfigure}[b]{0.47\textwidth}
         \centering
         \includegraphics[width=\textwidth]{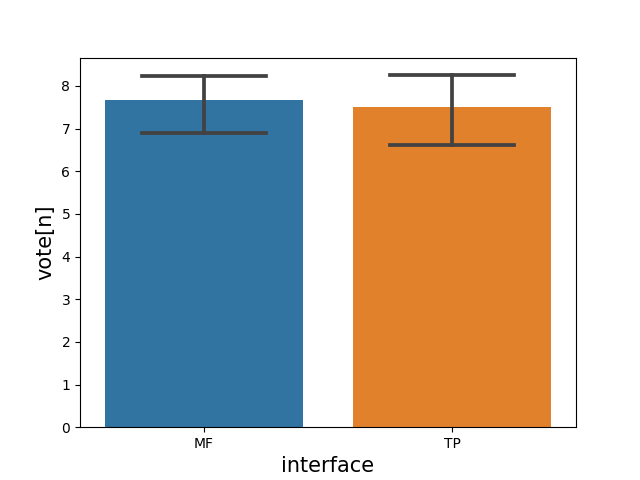}
         \caption{\textit{Interface intuitiveness, Table~\ref{tab:questionnaire} Q3.}}
         \label{fig:Task1_Int}
     \end{subfigure}
     \hfill
     \begin{subfigure}[b]{0.47\textwidth}
         \centering
         \includegraphics[width=\textwidth]{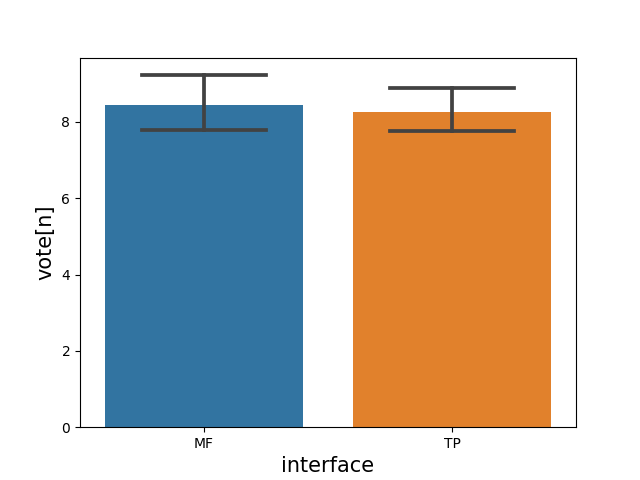}
         \caption{\textit{Interface speed, Table~\ref{tab:questionnaire} Q4.}}
         \label{fig:Task1_Vel}
     \end{subfigure}
        \caption{\textit{Task 1} experiments' results. \\ TP: UR10e teach pendant interface. MF: manipulation framework interface interface.}
        \label{fig:Task1_results_b}
\end{figure}

The experiments relative to Task 1 applied the method described in Section~\ref{ssec:experiments} and~\ref{sssec:task1_exp_desc}. Section \ref{ssec:progr_perf_meas} reports the parameters measured to evaluate the interfaces. All the evaluation indices measured during Task 1 are reported in Figures~\ref{fig:Task1_results_a} and~\ref{fig:Task1_results_b}.

The average learning time \textit{LeT} for the MFI is 44.9\% higher than the TPI, see Figure~\ref{fig:Task1_LeT}. This result was expected as the MFI has more complex concepts than the TPI and the learning time tends to be higher.  
The average programming time \textit{PrT} for the MFI is 51.7\% lower than the TPI, see Figure~\ref{fig:Task1_PrT}. High values of \textit{PrT} for the TPI are directly related to the number of objects involved in the task because the operator needs to teach multiple positions to perform collision-free trajectories. On the contrary, the MFI allows defining a \textit{Pick\&Place} \Task\ teaching only two grasping positions to define the \textit{Pick} \Action\ and two release positions to define the \textit{Place} \Action. Once defined as an \Action, the operator can replicate the same \Action.
The average number of programming questions \textit{PrQs} and test questions \textit{TeQs}, see Figures~\ref{fig:Task1_PrQ} and~\ref{fig:Task1_TeQ} respectively, is low for both the interfaces. These low values represent a good operator learning rate that reflects comparable ease of use for both interfaces.
The number of tests \textit{TeNs} and the test time \textit{TeT}, see Figure~\ref{fig:Task1_TeN} and~\ref{fig:Task1_TeT} respectively, are low. Most of the experiments do not present mistakes during the programming. The low presence of mistakes avoids corrections in many experiments; when required, the correction time is short, strengthening the result already given by \textit{PrT} values. The low number of mistakes leads to overlapping the \textit{TeTs} and the \textit{ExTs}.
The average execution time \textit{ExT} is 13.5\% lower for the TPI, see Figure~\ref{fig:Task1_ExT}. The robot motion planner interpolates the trajectory via-points taught through the TPI. The time to compute the trajectories is short, and the via-points are linearly interpolated. Instead, the MFI interpolates the starting and the goal position with optimal collision-free trajectories. The computational time to evaluate the planning scene and generates the collision-free trajectories can vary. This difference explains the differences in the execution times. Despite this, the difference between the results is insignificant; furthermore, TPI \textit{ExT} present a larger standard deviation than the MFI. The large standard deviation of the \textit{ExT} of TPI highlights a high dependency on the user's skills. The MFI presents a small \textit{ExT} standard deviation because the execution is independent of the operator's capacity.

In the end, a questionnaire was proposed to the users (see Table~\ref{tab:questionnaire}). The questions are related to the user's impressions concerning his/her previous experience level (Q1), ease of use (Q2), intuitiveness (Q3), learning speed (Q4), and the utility of his/her previous knowledge (Q5). Figure~\ref {fig:Task1_Comp},~\ref{fig:Task1_Int} and~\ref{fig:Task1_Vel} show almost identical results between the MFI and the TPI regarding ease of use, intuitiveness and learning speed. 
The plots of the score given to questions 1 and 5 (user's previous experience level and the utility of the previous knowledge to program the task) are not reported because the scores were always low. Therefore the informative contribution was not significant.
Table \ref{tab:users_info1_TP}, \ref{tab:users_info1_MF} shows information about the users that participated in the experiments of Task 1. 

\begin{table*}[tpb]
	\caption{Participants data for Task 1: E.Q. (educational qualification): h.s.g. (high school graduation); m.d. (master's degree);  b.d. (bachelor degree). I.R.P.E. (industrial robot programming experience). P.O. (professional occupation).} 
	\label{tab:Task1_hamp_results}
	\centering
	\footnotesize
	\subfloat[][ Group A]
	{
	    \begin{tabular}{cccccccc} 
		 \toprule
               & Age & E.Q.   & I.R.P.E.   & P.O.        & Gender \\
         \midrule
          1    & 24  & m.d.   & little     & PhD student & man    \\
          2    & 28  & m.d.   & no         & PhD student & man    \\
          3    & 35  & h.s.g. & little     & mechanical operator & man    \\
          4    & 23  & b.d.   & no         & student     & man    \\
          5    & 20  & h.s.g. & no         & student     & female  \\
          6    & 23  & m.d.   & little     & student     & man    \\
          7    & 26  & m.d.   & no         & PhD student & man    \\
          8    & 22  & h.s.g. & no         & student     & man    \\
          \bottomrule
        \end{tabular}
	\label{tab:users_info1_TP}
	}\\
	\vspace{5mm}
	\subfloat[][Group B]
	{
        \begin{tabular}{cccccccc} 
        \toprule
        & Age & E.Q.   & I.R.P.E. & P.O.               & Gender  \\
        \midrule
          9   & 29  & PhD    & expert     & researcher         & male     \\
          10  & 35  & h.s.g. & little     & mechanical operator               & male     \\
          11  & 23  & b.d.   & no       & student            & female   \\
          12  & 24  & b.d.   & little     & student            & male     \\
          13  & 24  & b.d.   & little     & student            & male     \\
          14  & 23  & b.d.   & no       & student            & female   \\
          15  & 22  & h.s.g. & no       & student            & male     \\
          16  & 31  & m.d.   & no       & sw developer & male     \\
          17  & 34  & m.d.   & no       & sw engineer   & male     \\
          \bottomrule
        \end{tabular}
		\label{tab:users_info1_MF}
		}
\end{table*}

\subsection{Task 2 Experiments Results}
\begin{figure}
     \centering
     \captionsetup{justification=centering}
     \begin{subfigure}[b]{0.46\textwidth}
         \centering
         \includegraphics[width=\textwidth]{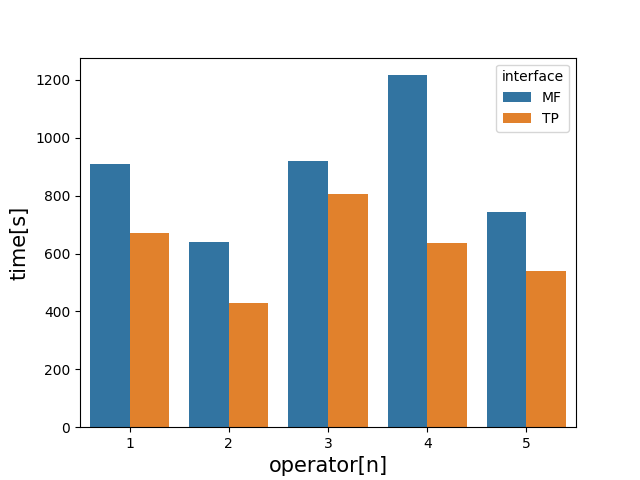}
         \caption{\textit{Learning Time (LeT).}}
         \label{fig:Task2_LeT}
     \end{subfigure}
     \hfill
     \begin{subfigure}[b]{0.46\textwidth}
         \centering
         \includegraphics[width=\textwidth]{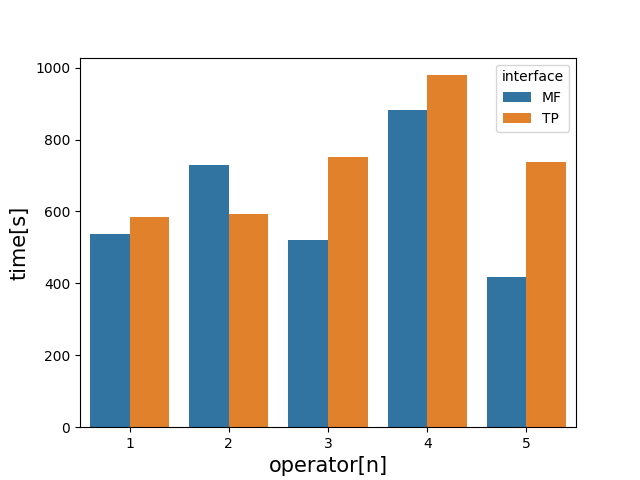}
         \caption{\textit{Programming Time (PrT).}}
         \label{fig:Task2_PrT}
     \end{subfigure}
     \hfill
     \begin{subfigure}[b]{0.46\textwidth}
         \centering
         \includegraphics[width=\textwidth]{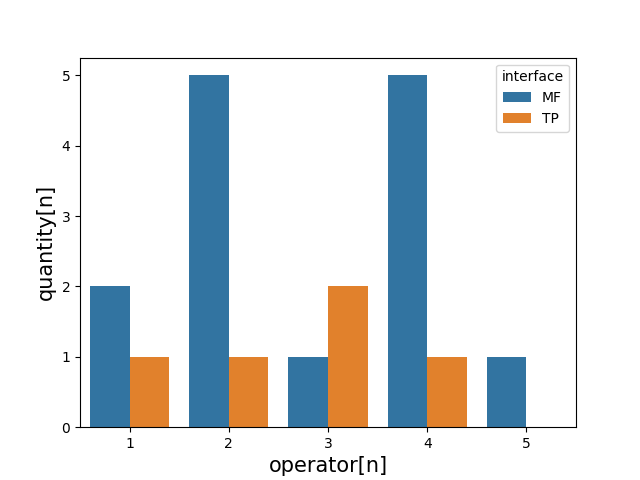}
         \caption{\textit{Programming Questions (PrQs).}}
         \label{fig:Task2_PrQ}
     \end{subfigure}
     \hfill
     \begin{subfigure}[b]{0.46\textwidth}
         \centering
         \includegraphics[width=\textwidth]{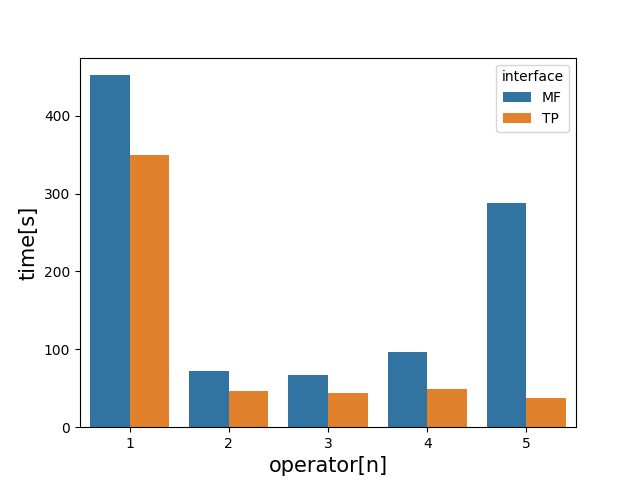}
         \caption{\textit{Test Time (TeT).}}
         \label{fig:Task2_TeT}
     \end{subfigure}
     \hfill
     \begin{subfigure}[b]{0.46\textwidth}
         \centering
         \includegraphics[width=\textwidth]{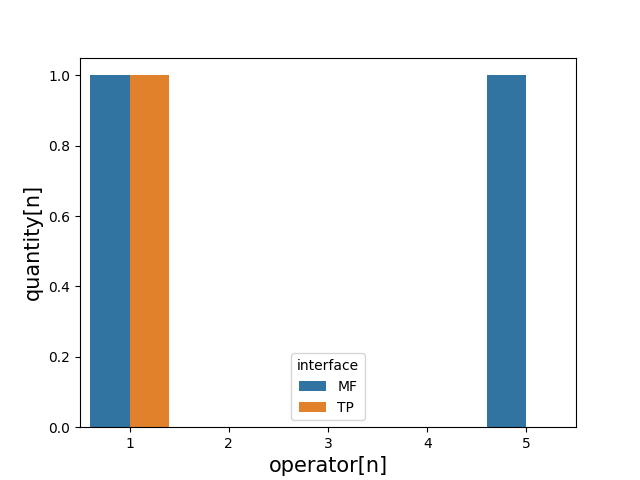}
         \caption{\textit{Test Numbers (TeNs).}}
         \label{fig:Task2_TeN}
     \end{subfigure}
     \hfill
     \begin{subfigure}[b]{0.46\textwidth}
         \centering
         \includegraphics[width=\textwidth]{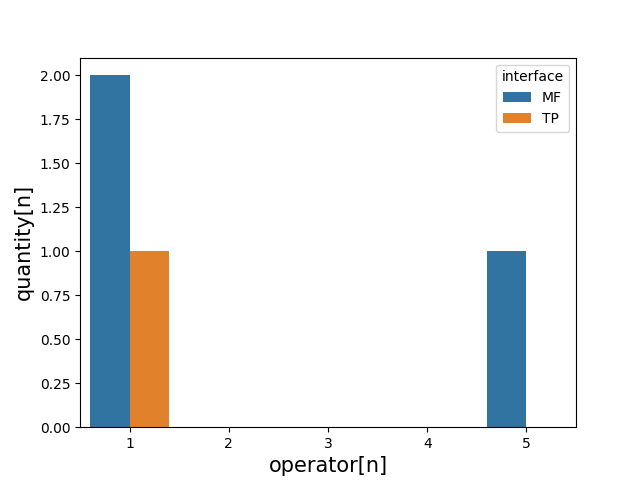}
         \caption{\textit{Test Questions (TeQs).}}
         \label{fig:Task2_TeQ}
     \end{subfigure}
     \hfill
     \begin{subfigure}[b]{0.46\textwidth}
         \centering
         \includegraphics[width=\textwidth]{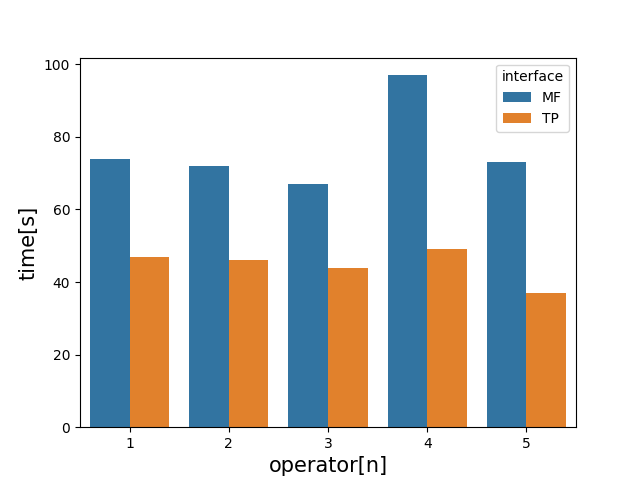}
         \caption{\textit{Execution Time (ExT).}}
         \label{fig:Task2_ExT}
     \end{subfigure}
     \hfill
     \begin{subfigure}[b]{0.46\textwidth}
         \centering
         \includegraphics[width=\textwidth]{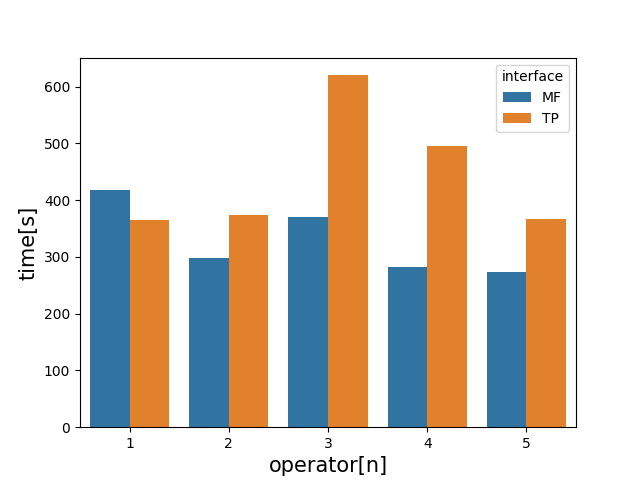}
         \caption{\textit{Reprogramming Time (ReT).}}
         \label{fig:Task2_ReT}
     \end{subfigure}
        \caption{\textit{Task 2} experiments' results. \\ TP: UR10e teach pendant. MF: manipulation framework interface.}
        \label{fig:Task2_results_a}
\end{figure}

\begin{figure}
     \centering
     \captionsetup{justification=centering}
     \begin{subfigure}[b]{0.46\textwidth}
         \centering
         \includegraphics[width=\textwidth]{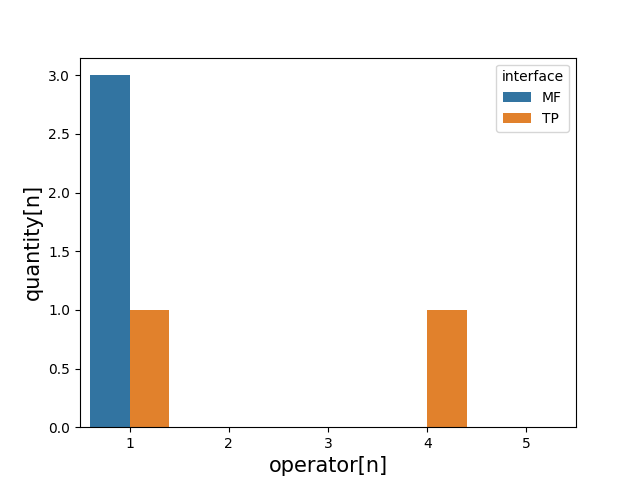}
         \caption{\textit{Reprogramming Questions (ReQs).}}
         \label{fig:Task2_ReQ}
     \end{subfigure}
     \hfill
     \begin{subfigure}[b]{0.46\textwidth}
         \centering
         \includegraphics[width=\textwidth]{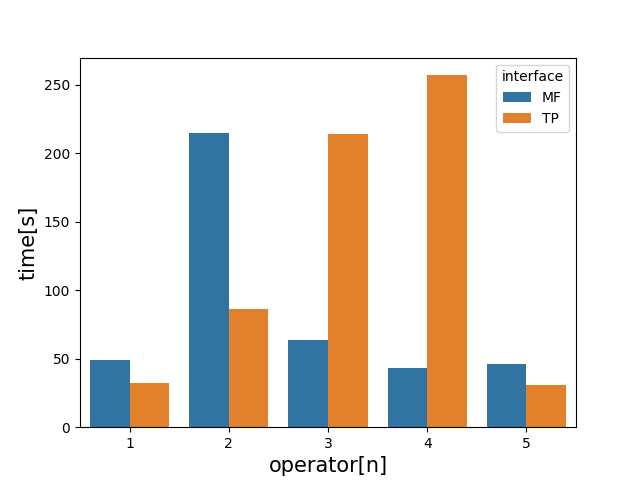}
         \caption{\textit{Reprogramming Test Time (ReTeT).}}
         \label{fig:Task2_ReTeT}
     \end{subfigure}
     \hfill
     \begin{subfigure}[b]{0.46\textwidth}
         \centering
         \includegraphics[width=\textwidth]{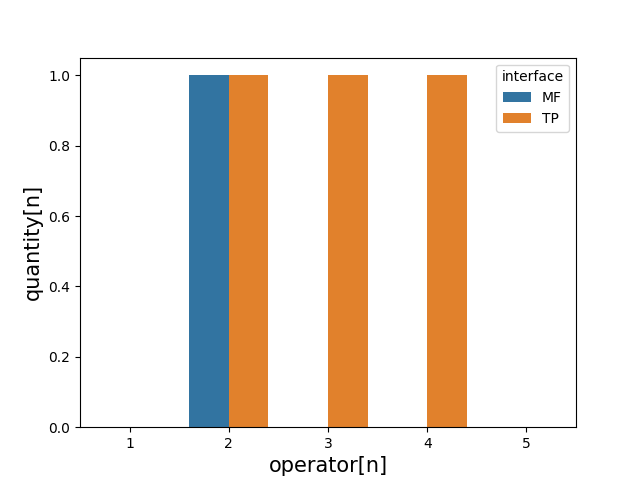}
         \captionsetup{width=1.1\linewidth}
         \caption{\textit{Reprogramming Test Numbers (ReTeN).}}
         \label{fig:Task2_ReTeN}
     \end{subfigure}
     \hfill
     \begin{subfigure}[b]{0.46\textwidth}
         \centering
         \includegraphics[width=\textwidth]{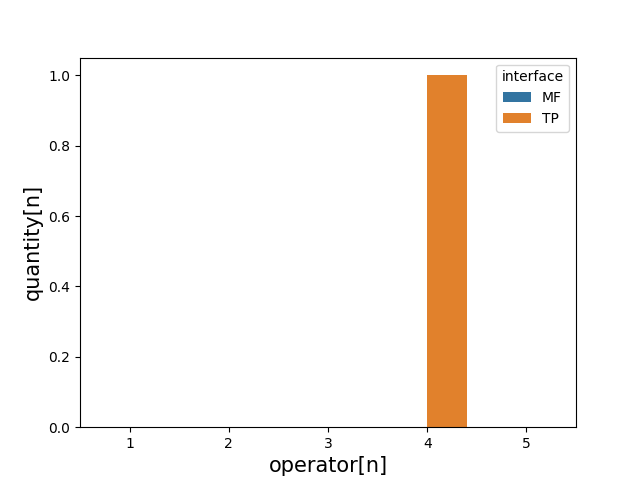}
         \captionsetup{width=1.1\linewidth}
         \caption{\textit{Reprogramming Test Questions (ReTeQ).}}
         \label{fig:Task2_ReTeQ}
     \end{subfigure}
     \hfill
     \begin{subfigure}[b]{0.46\textwidth}
         \centering
         \includegraphics[width=\textwidth]{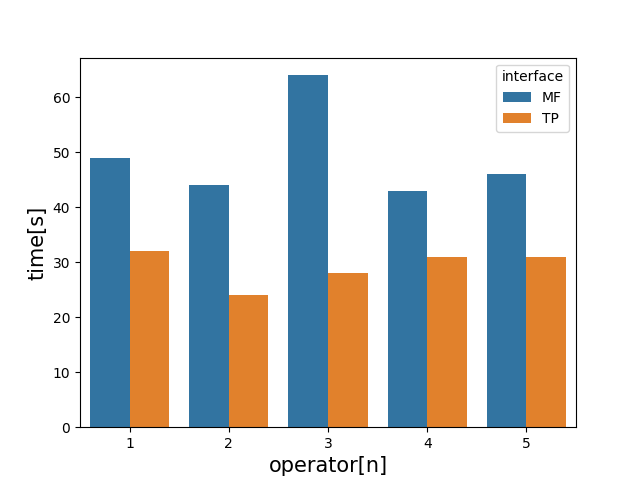}
         \captionsetup{width=1.1\linewidth}
         \caption{\textit{Reprogramming Execution Time (ReExT).}}
         \label{fig:Task2_ReExT}
     \end{subfigure}
     \hfill
     \begin{subfigure}[b]{0.46\textwidth}
         \centering
         \includegraphics[width=\textwidth]{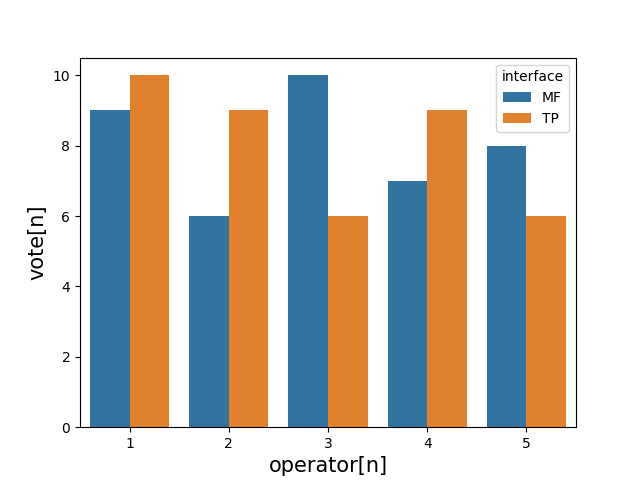}
         \caption{\textit{Interface ease of use, Table~\ref{tab:questionnaire} Q2.}}
         \label{fig:Task2_Comp}
     \end{subfigure}
     \hfill
     \begin{subfigure}[b]{0.46\textwidth}
         \centering
         \includegraphics[width=\textwidth]{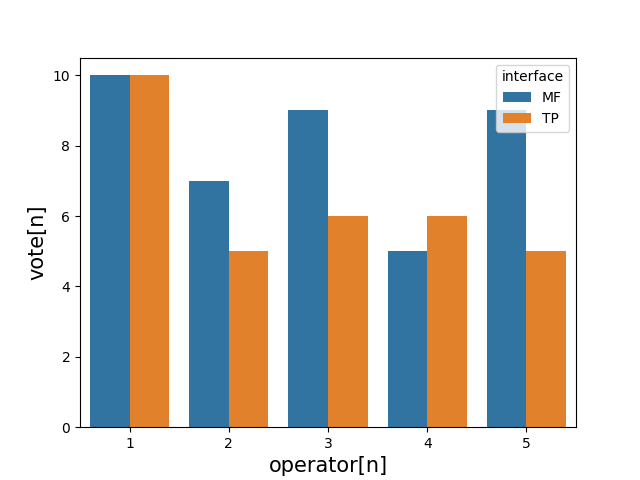}
         \caption{\textit{Interface intuitiveness, Table~\ref{tab:questionnaire} Q3.}}
         \label{fig:Task2_Int}
     \end{subfigure}
     \hfill
     \begin{subfigure}[b]{0.46\textwidth}
         \centering
         \includegraphics[width=\textwidth]{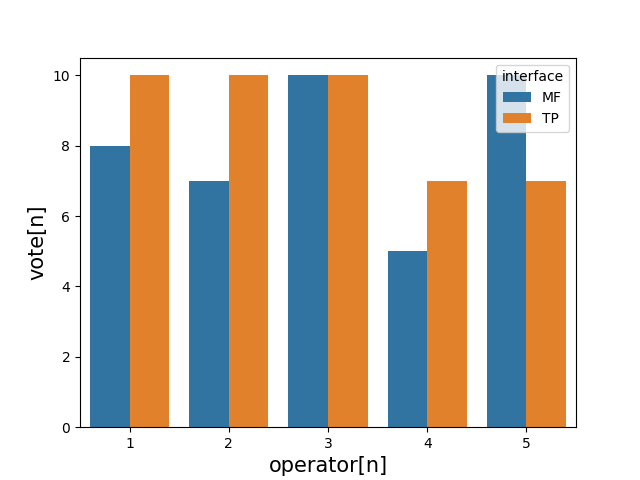}
         \caption{\textit{Interface speed, Table~\ref{tab:questionnaire} Q4.}}
         \label{fig:Task2_Vel}
     \end{subfigure}
        \caption{\textit{Task 2} experiments' results. \\ TP: UR10e teach pendant. MF: manipulation framework interface.}
        \label{fig:Task2_results_b}
\end{figure}

The experiments relative to \textit{Task 2} applied the method described in Section~\ref{ssec:experiments} and ~\ref{sssec:task2_exp_desc}. Section~\ref{ssec:progr_perf_meas} reports the parameters measured to evaluate the interfaces. All the evaluation indices measured during Task 2 are reported in Figures~\ref{fig:Task2_results_a} and~\ref{fig:Task2_results_b}.

The average learning time \textit{LeT} is 43.7\% higher for MFI; see Figure~\ref{fig:Task2_LeT}. The result is similar to the \textit{LeT} of Task 1 experiments.
The average programming time \textit{PrT} is 15.2\% lower for the MFI; see Figure~\ref{fig:Task2_PrT}. The \textit{PrT} is similar for both the interfaces because \textit{Task 2} is composed of only two objects, and the number of repetitive \Actions\ is reduced, tending to provide similar results. 
The programming questions \textit{PrQs} is higher for the MFI; see Figure~\ref{fig:Task2_PrQ}. The users involved in Task 2 (\ie shop floor machine tools operators and technicians) had less familiarity with the use of robots and, in general, less inclination to technologies compared to the users of \textit{Task 1} (\ie STEM faculties students). This aspect is the possible cause of why \textit{Task 2} \textit{PrQs} values are higher than \textit{Task 1}, in particular for the MFI. The more complex structure of MFI has amplified this phenomenon. Despite this problem, the \textit{PrQs} values do not represent a real problem. The training of non-expert operators for the MFI is less than one hour (considering the video watching). 
The test time \textit{TeT}, see Figure~\ref{fig:Task2_TeT}, shows similar \textit{TeT} for both interfaces apart from the spike of user one that is however, present for both interfaces; in particular, the time to correct the errors is comparable.
The test numbers \textit{TeNs}, see Figure~\ref{fig:Task2_TeN}, show a not relevant number of trials, so the reduced number of errors made during the programming demonstrates that reduced knowledge does not affect the operator performance.
The test questions \textit{TeQs}, see Figure~\ref{fig:Task2_TeQ}, show a high autonomy of the user to correct the errors. 
The execution time \textit{ExT}, see Figure~\ref{fig:Task2_ExT}, indicates that the TPI \textit{ExT} has lower values than the MFI. In this case, the computation time necessary for the MF to generate collision-free trajectories is higher than Task 1 because the robot workspace presents constrained spaces and more obstacles. The higher \textit{ExTs} is reflected in the benefit of a collision-free trajectory guaranteed by the MF motion planners. On the contrary, with the TPI, the collision avoidance of the robot is in charge of the programmer.
The average reprogramming time \textit{ReT}, see Figure~\ref{fig:Task2_ReT}, shows that usually, the MFI \textit{ReT} values are lower than the TPI. The MFI \textit{ReT} average value is 26.1\% lower than TPI. With the MFI, the user has to modify the program to teach only two new positions. Instead, the TPI requires adding new via-points to the trajectories already defined.
The reprogramming number of questions \textit{ReQs}, see Figure~\ref{fig:Task2_ReQ}, shows the low \textit{ReQs} values; most of the users did not need any help during the reprogramming. 
The reprogramming test times \textit{ReTeTs}, the reprogramming tests numbers \textit{ReTeN}, the test questions \textit{ReTeQs}, and the execution time \textit{ReExT}, see Figure~\ref{fig:Task2_ReTeT}, \ref{fig:Task2_ReTeN}, \ref{fig:Task2_ReTeQ}, and \ref{fig:Task2_ReExT} respectively, show results similar to those obtained in the first testing phase and no significant differences emerged between the interfaces. 

As for Task 1, at the end of Task 2, the questionnaire described in Table~\ref{tab:questionnaire} was proposed to the users. Figure \ref{fig:Task2_Comp}, \ref{fig:Task2_Int}, and \ref{fig:Task2_Vel} show again similar values for both interfaces. The parameters have high values demonstrating a good appreciation by the machine tools operators highlighting the usability of the interfaces in the industrial world. The assessments of previous knowledge have no informative contribution, and the plot of the scores of questions 1 and 5 are not reported.
Table~\ref{tab:users_info2} show the information about the users participating in the experiments of Task 2. 

\begin{table}[tpb]
	\caption{Participants data for Task 2: E.Q. (educational qualification): h.s.g. (high school graduation); m.d. (master's degree);  b.d. (bachelor degree). I.R.P.E. (industrial robot programming experience). P.O. (professional occupation).} 
	\centering
	\footnotesize
        \begin{tabular}{cccccccc} 
        \toprule
              & Age & E.Q.   & I.R.P.E. & P.O.               & Gender  \\
        \midrule
          1   & 36  & h.s.g. & little     & mechanical operator & male    \\
          2   & 35  & h.s.g. & little     & mechanical operator & male    \\
          3   & 32  & h.s.g. & No       & mechanical operator & male    \\
          4   & 46  & h.s.g. & No       & shift supervisor & male    \\
          5   & 26  & m.d.   & little     & process engineer & male    \\
          \bottomrule
        \end{tabular}
		\label{tab:users_info2}
\end{table}

\subsection{Discussion}

The experiments presented similar results for both interfaces. The similarities are visible in every operational aspect of the interfaces. In other words, the MFI can compete with the UR10e TPI. The higher learning times of the MFI can be quickly recovered by the daily use of the MF for robot programming when the tasks present repetitive \Actions.
However, the learning time of the MFI for a non-expert end-user was minimal and below one hour to give a basic understanding of the interface and to be independent of the task programming. At the same time, the MF brings together several additional benefits, such as using the ROS framework, \textit{task-oriented} programming, and collision-free motion planning. 
The goal of using a complex robot programming framework accompanied by an intuitive interface was reached, demonstrating that shop floor operators can quickly adopt the framework without significant learning barriers and any particular preliminary knowledge. 
This paper does not present an alternative to UR10e TP. The paper aims to demonstrate that using complex software with a suitable programming interface can bring advanced programming features even to shop floor operators, enabling the use of advanced robotic cells in SMEs. 

\section{Conclusions and Future Works}\label{sec:conc_and_fut_works}

This paper compares two robot programming interfaces representing two different robot programming approaches. The first is the UR10e teach pendant interface, combined with \textit{lead-through} programming, enabling a classic \textit{robot-oriented} approach. The second is the Manipulation Framework interface providing \textit{task-oriented} programming approach. 
The experiments over 22 users show similar results for both interfaces highlighting that an intuitive interface that hides the complexity of a framework based on ROS, such as the MF, can reach a high level of acceptance between end-users without specific programming experience. The short learning times show the possibility of training an end-user in very little time bringing advanced features to the shop floor without particular knowledge of robotics. At the same time, the flexibility during reprogramming improved. The modification of the robot control program can be accessible to all production plant operators.

\textcolor{red}{In future work, the authors will investigate using machine learning algorithms to create new actions. From a limited number of skills, for example, machine learning algorithms should dynamically compose new actions based on specific requirements without developer interventions.}

\section*{Acknowledgement}
This work is partially supported by ShareWork project (H2020, European Commission – G.A. 820807).

\noindent The authors thank Cembre S.p.A.~\citep{Cembre:2022} to provide the opportunity to make the experimental tests on their shop floor with their machine tool operators. A special thanks to Andrea Scala and Piervincenzo Tavormina for help organizing the experiments.

 \bibliographystyle{tfcad} 
 \bibliography{mybib}

\end{document}